\documentclass[letterpaper, 10 pt, conference]{ieeeconf}  

\IEEEoverridecommandlockouts                              

\usepackage{graphicx}
\usepackage{threeparttable}
\usepackage{booktabs}
\usepackage{amsmath} 
\usepackage{amssymb}  

\title{\LARGE \bf
Cross-Subject Semantic Decoding with Shared-Space Alignment for Generalized Neural Representation Learning
}

\author{Ji-Hoon Heo$^{1}$, Aleksandra Wisniewska$^{1}$, Seo-Hyun Lee$^{2}$, and Seong-Whan Lee$^{1}$, \textit{Fellow, IEEE}
\thanks{This work was partly supported by Institute of Information \& Communications Technology Planning \& Evaluation (IITP) grant funded by the Korea government (MSIT) (No. RS-2019-II190079, Artificial Intelligence Graduate School Program (Korea University), and No. RS-2024-00336673, AI Technology for Interactive Communication of Language Impaired Individuals).}%
\thanks{$^{1}$ Ji-Hoon Heo, Aleksandra Wisniewska, and Seong-Whan Lee are with the Department of Artificial Intelligence, Korea University, Seoul, Republic of Korea
        {\tt\small jihoon\_heo@korea.ac.kr, awisniewska@korea.ac.kr, sw.lee@korea.ac.kr}}%
\thanks{$^{2}$Seo-Hyun Lee is with the Department of Brain and Cognitive Engineering, Korea University, Seoul, Republic of Korea
        {\tt\small seohyunlee@korea.ac.kr}}%
}

\begin{document}

\maketitle
\thispagestyle{empty}
\pagestyle{empty}

\begin{abstract}

Generalizing across subjects remains challenging in invasive neural recordings because electrode configurations, anatomical structures, and neural signal patterns vary substantially across individuals. To investigate such inter-subject variability, we propose a cross-subject semantic decoding framework that aligns neural responses to speech perception from multiple subjects into a shared latent space and learns a mapping from the aligned neural representations to contextual embeddings. More specifically, using electrocorticography data collected during natural language comprehension, we estimate the shared space using the shared response model and train a decoder to predict contextual semantic embeddings from projected neural responses. For a held-out subject, we estimate a subject-specific projection into the predefined shared space, and directly apply the pretrained decoder without any retraining. Experimental results demonstrate that the proposed framework consistently outperforms baseline methods across evaluation settings and exhibits a reduced performance drop from source subject to held-out subject testing, indicating improved cross-subject generalization. These results suggest that aligning neural activity into a shared latent space, while decoding in a semantic embedding space, provides an effective strategy for improving cross-subject generalization by reducing subject-specific differences in neural responses while effectively capturing shared stimulus-related representations.

\end{abstract}

\section{INTRODUCTION}

Over the past decades, advances in artificial intelligence and deep learning have demonstrated promising generalization performance across diverse domains, including computer vision, natural language processing, speech processing, and biomedical signal analysis \cite{kim2022automatic, zhao2020diagnosis, yu2019weighted, shi2019leveraging, lee2025hierspeech++}. Recent studies demonstrated that models can generalize to unseen data distributions by learning representations that capture shared underlying structures across domains \cite{ganin2016domain, muandet2013domain}. However, in speech brain--computer interfaces (BCIs), achieving comparable generalization across subjects remains challenging due to high inter-subject variability and limited availability of task-specific neural data \cite{chen2024neural, singh2025transfer}.

Speech BCIs aim to decode linguistic information from neural signals to support communication, particularly for individuals with severe motor or speech impairments. Recent advances have shown potential for robust speech decoding and imagined speech-based communication systems \cite{moses2021neuroprosthesis, lee2023towards, leedecoding2021, willett2023high, card2024accurate, lee2022toward}. However, most existing approaches remain subject-dependent because neural signals vary substantially across subjects due to differences in electrode placement, electrode count, cortical structure, and neural response patterns \cite{silva2024speech}, making it difficult to learn shared neural representations even when presented with the same linguistic stimuli. Moreover, collecting large-scale neural data over extended periods is challenging in clinical settings due to factors such as patient condition and fatigue \cite{jeong2020multimodal}. These limitations highlight the need for models that generalize to new users while maintaining stable performance with limited data.

Recent studies have demonstrated the potential of neural response hyperalignment \cite{chen2015reduced}, which can be applied to electrocorticography (ECoG) data to alleviate inter-subject variability in neural response representations \cite{bhattacharjee2026aligning}. This approach can resolve the mismatch in input dimensionality arising from differences in electrode counts across subjects, and effectively extract shared neural features embedded in the latent space \cite{chen2015reduced}. Based on these observations, we hypothesize that rather than adopting a conventional encoding model that predicts neural responses from text embeddings, training a decoding model that predicts corresponding text embeddings from shared neural features would facilitate more stable performance on held-out subjects by projecting their neural responses into the shared space.

\begin{figure*}
    \centering
    \includegraphics[width=\linewidth]{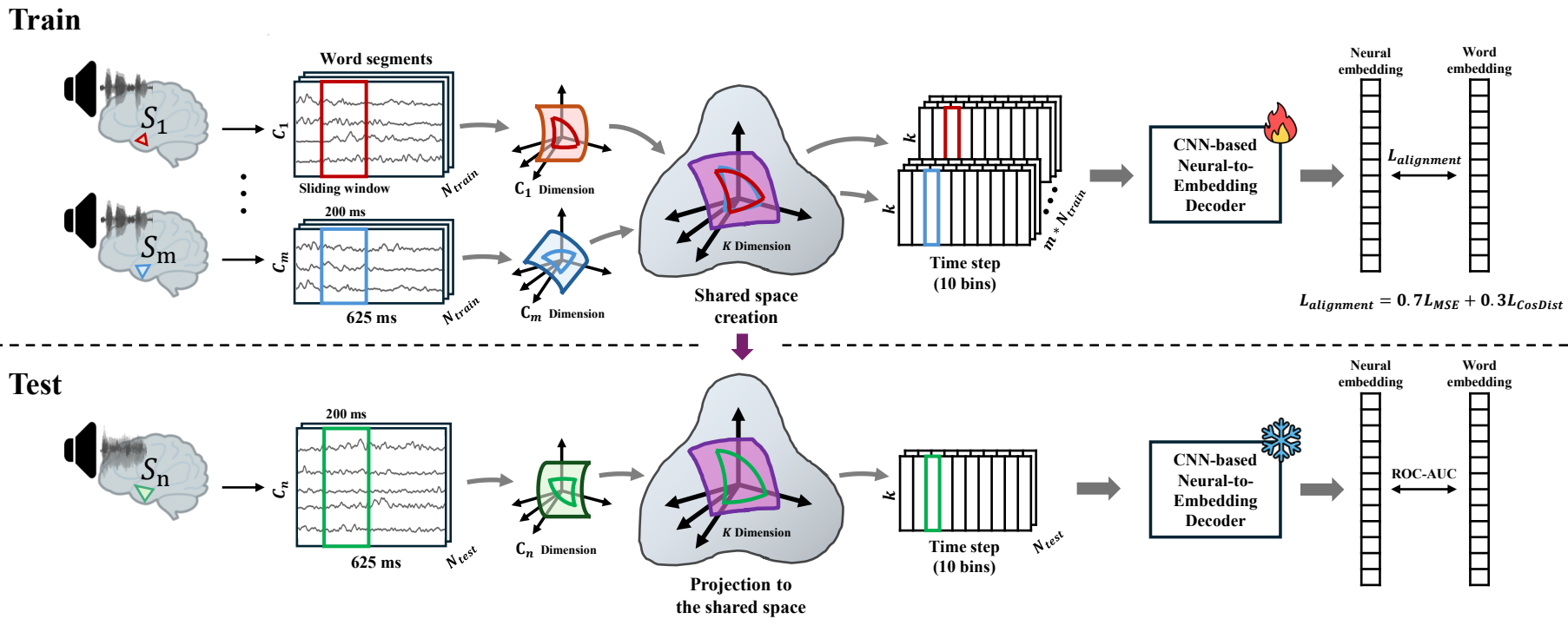}
    \caption{\textbf{Overview of the proposed cross-subject semantic decoding framework.}
    \textbf{Train (top):} Neural response signals from multiple source subjects are segmented into word-aligned windows (625 ms) and further divided into overlapping temporal bins using a sliding-window approach (200 ms). For each temporal bin, time-averaged neural responses are used to estimate a bin-specific shared latent space via the shared response model (SRM). Each time-bin-specific neural response is then projected into its corresponding shared space, and the resulting shared representations are concatenated along the temporal dimension to construct a structured representation. A CNN-based neural-to-embedding decoder is trained to map the shared neural representations to corresponding contextual semantic embeddings.
    \textbf{Test (bottom):} For a held-out subject, a subject-specific projection matrix is estimated to align its neural response signals to the pre-defined shared space without retraining the decoder. The projected temporal representations are fed into the pretrained decoder to predict semantic embeddings. Decoding performance is evaluated in the semantic embedding space by comparing predicted and ground-truth embeddings.}
    \label{fig:main}
\end{figure*}

In this context, an important question is how to extract and align shared features across subjects to enable effective generalization. Prior speech decoding studies have primarily focused on extracting phonological features or decoding low-level linguistic representations from neural signals \cite{meng2022evidence, chen2024neural, ahn2022multiscale}. Although these approaches have the advantage of being closely linked to audible speech signals, they may also more strongly reflect subject-specific variability in neural responses \cite{arazi2019neural}. In this regard, recent studies have shown that high-level linguistic representations, particularly contextual embeddings, explain neural responses during language processing more effectively than lower-level representations \cite{zada2025podcast}. In addition, contextual embedding spaces derived from language models have been reported to share structural properties with neural representations \cite{goldstein2022shared}.

In this study, we present a framework to evaluate the generalization capability of neural decoding models in predicting semantic embeddings from neural responses using the shared response model (SRM)~\cite{chen2015reduced}. Based on the previous findings~\cite{bhattacharjee2026aligning, goldstein2022shared}, we adopted an approach that predicts contextual embeddings for each word, with the aim of focusing on relatively consistent semantic structures across subjects rather than low-level response characteristics that may vary across individuals. Rather than following prior SRM-based approaches that focus on learning encoding models within a shared space, we employ SRM to train a decoding model on shared neural features and evaluate its ability to generalize to held-out subjects by estimating only subject-specific projections. This approach shifts the focus from subject-specific modeling to shared representation learning, enabling more efficient generalization to held-out subjects and suggesting a scalable direction for developing universal neural decoding models.

\section{Materials and Methods}

\subsection{Dataset Description}

We analyzed a public ECoG dataset by Zada et al., \cite{zada2025podcast}. Neural recordings were collected from nine participants who underwent clinical intracranial electrode implantation for epilepsy treatment. The dataset includes a total of 1,330 electrodes, with electrode placements varying across participants depending on clinical requirements. During the experiment, participants listened to a naturalistic spoken audio (podcast) of approximately 30 minutes, containing over 5,000 individual words. The audio was manually transcribed, and precise temporal alignment (onset and offset) was provided for each word. Neural signals were preprocessed and standardized to a sampling rate of 512 Hz. In addition, complementary linguistic features aligned at the word level are provided, including acoustic features, phoneme-level representations, syntactic information, non-contextual embeddings, and contextual embeddings derived from large language models.

High-gamma power (70--200 Hz) was extracted from neural signals using a 6-cycle wavelet transform, consistent with prior work~\cite{goldstein2022shared}. Although prior alignment work reported 184 electrodes~\cite{bhattacharjee2026aligning}, the associated public podcast benchmark repository provides a list of only 183 electrodes, which we used for consistency with the released implementation and reproducibility.
For semantic alignment, we selected 5,136 words from the dataset and used contextual embeddings, which were reported to achieve the highest alignment performance with neural responses~\cite{goldstein2022shared,bhattacharjee2026aligning}, as the corresponding semantic representations. These embeddings were used as the target space corresponding to neural representations, and the model was designed to learn a mapping from the shared neural space to the contextual embedding space.

\subsection{Decoding Framework}
The overall pipeline of the proposed framework is illustrated in Fig.~1, which provides a schematic overview of the data processing, shared-space generation and application, and decoding stages.

\subsubsection{Data Processing}
To perform word-level analysis from raw neural response signals, we applied epoching for each stimulus word. The middle time step for each word segment was lagged by +150 ms relative to the word onset, following prior work that reported the best decoding performance at this lag \cite{goldstein2022shared}. Each segment was constructed with a total duration of 625 ms, allowing us to extract neural responses corresponding to each word within a fixed temporal window.

For each subject $i$, the data are represented as
\begin{equation}
X_i \in \mathbb{R}^{C_i \times N \times T_{\text{raw}}},
\end{equation}
where $C_i$ denotes the number of electrodes for subject $i$, $N$ is the number of words, and $T_{\text{raw}}$ is the number of raw temporal samples within each word segment.

In conventional SRM-based alignment, each segment is averaged along the temporal dimension to obtain
\begin{equation}
\bar{X}_i \in \mathbb{R}^{C_i \times N},
\end{equation}
which is then used for shared space estimation. However, this process removes all temporal information within each segment. To address this limitation, we divide each \mbox{625 ms} segment into sliding windows of 200 ms and perform temporal averaging within each window, resulting in $T = 10$ temporal bins. For each time bin $t$, the data are represented as
\begin{equation}
X_i^{(t)} \in \mathbb{R}^{C_i \times N}, \quad t = 1, \dots, T.
\end{equation}
We then apply SRM independently to each time bin, estimating time-specific shared representations and projection matrices for each temporal segment.

\subsubsection{Applying Shared Response Model}

Prior to applying the SRM \cite{chen2015reduced}, we define all participants except the held-out subject as source subjects. For each source subject $i$, the train data at time bin $t$ are represented as $X_{i, train}^{(t)} \in \mathbb{R}^{C_i \times N_{\text{train}}}$, where $N_{\text{train}}$ denotes the number of training words. SRM jointly estimates a shared representation $S^{(t)} \in \mathbb{R}^{k \times N_{\text{train}}}$ and subject-specific projection matrices $W_i^{(t)} \in \mathbb{R}^{C_i \times k}$ to align all source subjects into a common latent space. This is formulated as the following optimization problem:
\begin{equation}
\begin{aligned}
\min_{W_i^{(t)},\, S^{(t)}} \quad & \sum_i \left\| X_{i,\text{train}}^{(t)} - W_i^{(t)} S^{(t)} \right\|_F^2 \\
\text{s.t.} \quad & {W_i^{(t)}}^\top W_i^{(t)} = I.
\end{aligned}
\end{equation}

Here, $S^{(t)}$ represents the shared neural representation across source subjects, and $W_i^{(t)}$ is a subject-specific basis matrix that maps the $k$-dimensional shared space to the \mbox{$C_i$-dimensional} electrode space of subject $i$. After optimization via SRM, each source subject’s data are transformed as
\begin{equation}
Z_{i,train}^{(t)} = {W_i^{(t)}}^\top X_{i,train}^{(t)} \in \mathbb{R}^{k \times N_{\text{train}}},
\end{equation}
resulting in a shared neural representation. The representations $Z_i^{(t)}$ obtained from different time bins are then concatenated along the temporal axis to construct a $k \times T$ representation for each word. These are further concatenated across all $M$ source subjects along the sample axis, yielding a final representation $Z_{train} \in \mathbb{R}^{(N_{\text{train}} \times M) \times k \times T}$, which is used as input for training the decoding model.

For the held-out subject, the shared representation $S^{(t)}$ learned from the source subjects is fixed, and only the subject-specific projection matrix is estimated to align the data into the same shared space. This can be viewed as a special case of Eq.~(4), where the shared representation is fixed and the optimization is performed only over a single subject. Specifically, given the training data \mbox{$X_{\text{target}}^{(t)} \in \mathbb{R}^{C_{\text{target}} \times N_{\text{train}}}$} aligned to the same stimulus and sample order as the source-subject training data, the projection matrix $W_{\text{target}}^{(t)}$ is obtained by minimizing the same objective in Eq.~(4) with respect to $W_{\text{target}}^{(t)}$ while keeping $S^{(t)}$ fixed.

During the test, all subjects, including the held-out subject, are evaluated individually. The test data at time bin $t$ of each subject are projected into the shared space using the corresponding subject-specific transformation as
\begin{equation}
\begin{aligned}
Z_{i,\text{test}}^{(t)} &= {W_i^{(t)}}^\top X_{i,\text{test}}^{(t)} \in \mathbb{R}^{k \times N_{\text{test}}}, \\
& \forall i \in \{\text{all subjects}\}.
\end{aligned}
\end{equation}
After projection, the test data at time bin $t$ are concatenated along the temporal axis to construct a final test representation $Z_{i, test} \in \mathbb{R}^{N_{\text{test}} \times k \times T}$ for each subject.
As a result, the test data from all subjects are aligned into a common shared space, and the resulting shared neural representations are used to evaluate the trained decoding model.

\begin{figure}
    \centering
    \includegraphics[width=0.5\textwidth]{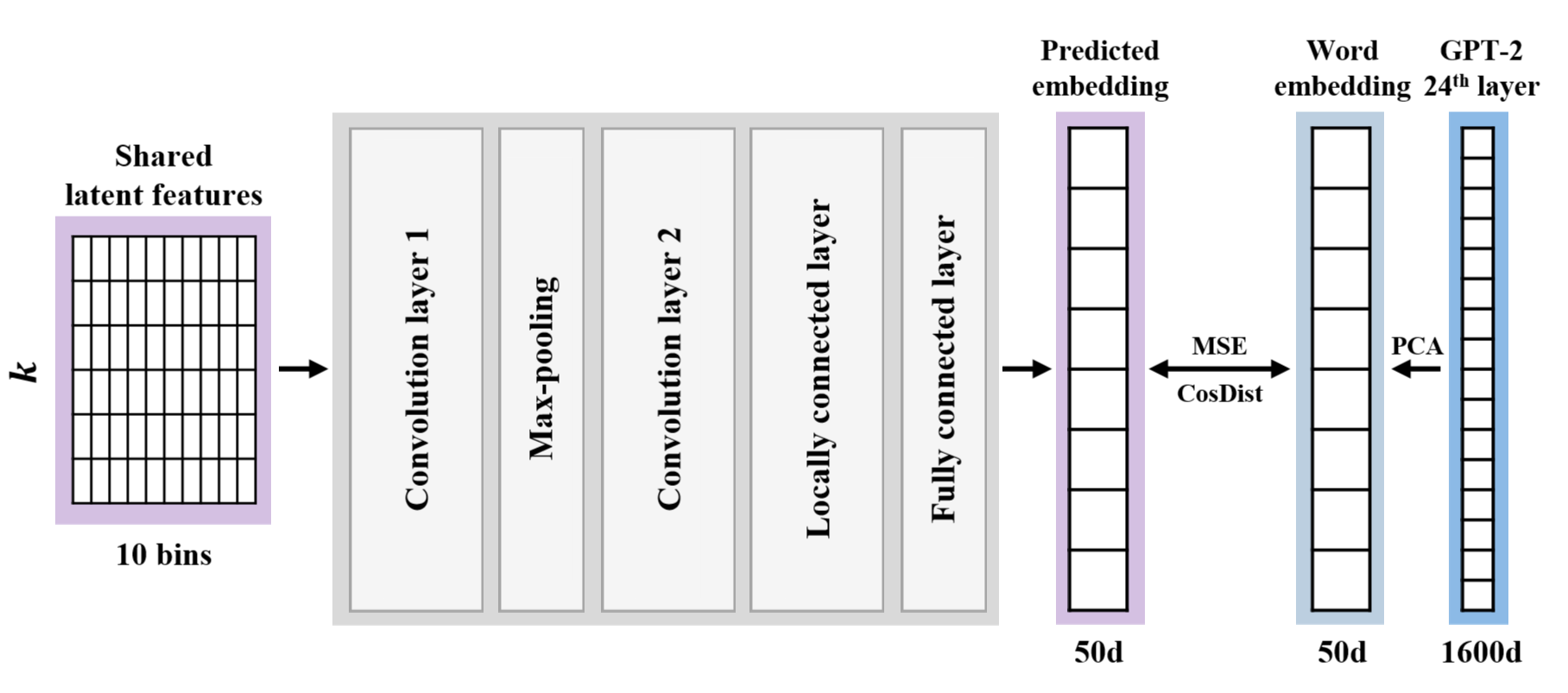}
    \caption{\textbf{Decoding model architecture.} The model takes the shared latent neural representation $Z$ as input and predicts the corresponding semantic embedding $\hat{y}$. The input, represented as a matrix in the shared space, is passed through a convolutional neural network (CNN). The model is trained to map subject-aligned neural activity to its semantic representation by minimizing a combination of mean squared error (MSE) and cosine distance (CosDist) between the predicted embedding and the ground-truth semantic embedding. Because all neural inputs are projected into a shared space prior to decoding, the model learns a unified mapping across subjects using pooled source-subject data.}
    \label{fig:model}
\end{figure}

\subsubsection{Neural Decoding Model}

The decoding model is based on prior work that aligns neural response signals with word embeddings \cite{goldstein2022shared}, and the overall architecture is illustrated in Fig.~2. The model takes shared neural response representations in the common latent space as input and predicts the corresponding target word embeddings using a deep convolutional neural network. The target word embeddings are contextual embeddings obtained from the 24th layer of the GPT-2 language model \cite{radford2019language}, with dimensionality reduced using PCA (k=50). The model is trained to minimize both the mean squared error and cosine distance between the predicted and ground-truth embeddings, thereby learning a mapping that aligns shared neural response patterns with the semantic embedding space.

All our models were trained under a consistent set of hyperparameters across all held-out subjects and dimensionality reduction methods. We used the AdamW optimizer with a learning rate of $5 \times 10^{-4}$ and weight decay of $1 \times 10^{-4}$, and trained the model with a batch size of 32 for up to 100 epochs. Early stopping was applied based on validation cosine similarity with a patience of 10 epochs. The training objective was a weighted combination of mean squared error and cosine distance, with empirically determined weights of 0.7 and 0.3, respectively.

\subsection{Experimental Design}

\subsubsection{Source and Target Configuration}

We evaluate the proposed framework in a held-out-subject setting, where the target subject is excluded from decoder training and only a subject-specific projection is estimated using the target subject’s training data. This setting allows us to explicitly evaluate whether the learned shared representation supports generalization to held-out subjects without updating model parameters.

\begin{table}[t]
\centering
\caption{Source-subject decoding performance.}
\label{tab:src_results}
\setlength{\tabcolsep}{6pt}
\renewcommand{\arraystretch}{1.1}
\begin{tabular}{lccc}
\toprule
\textbf{Metric} & \textbf{SRM Src.} & \textbf{PCA+HA Src.} & \textbf{PCA Src.} \\
\midrule
AUC-ROC       & $\mathbf{0.671 \pm 0.007}$ & $0.656 \pm 0.006$ & $0.637 \pm 0.005$ \\
Pairwise Acc. & $\mathbf{0.667 \pm 0.007}$ & $0.656 \pm 0.005$ & $0.634 \pm 0.005$ \\
Top-10        & $\mathbf{0.055 \pm 0.004}$ & $0.050 \pm 0.002$ & $0.043 \pm 0.002$ \\
Top-50        & $\mathbf{0.222 \pm 0.007}$ & $0.209 \pm 0.006$ & $0.187 \pm 0.005$ \\
\bottomrule
\end{tabular}
\begin{tablenotes}[flushleft]
\footnotesize
\item Src.: Source-subject evaluation; Acc.: Accuracy.
\end{tablenotes}
\end{table}

\begin{table}[t]
\centering
\caption{Cross-subject (held-out) decoding performance.}
\label{tab:ho_results}
\setlength{\tabcolsep}{6pt}
\renewcommand{\arraystretch}{1.1}
\begin{tabular}{lccc}
\toprule
\textbf{Metric} & \textbf{SRM H/O} & \textbf{PCA+HA H/O} & \textbf{PCA H/O} \\
\midrule
AUC-ROC       & $\mathbf{0.662 \pm 0.044}$ & $0.640 \pm 0.040$ & $0.553 \pm 0.029$ \\
Pairwise Acc. & $\mathbf{0.656 \pm 0.035}$ & $0.638 \pm 0.036$ & $0.549 \pm 0.032$ \\
Top-10        & $\mathbf{0.051 \pm 0.016}$ & $0.044 \pm 0.012$ & $0.023 \pm 0.006$ \\
Top-50        & $\mathbf{0.209 \pm 0.039}$ & $0.193 \pm 0.033$ & $0.119 \pm 0.020$ \\
\bottomrule
\end{tabular}
\begin{tablenotes}[flushleft]
\footnotesize
\item H/O: Held-out subject evaluation; Acc.: Accuracy.
\end{tablenotes}
\end{table}

\subsubsection{Data Splitting and Evaluation Procedure}

To evaluate model performance, the dataset is partitioned into five non-overlapping sequential splits. Each split is iteratively selected as the test set, while the remaining four splits are used for training. This process is repeated five times, and the final performance for each subject is obtained by averaging across the five splits. Within each split, the following procedure is applied. First, SRM (or baseline alignment) is estimated using the training data of the source subjects. Second, the decoding model is trained on the aligned representations of the source subjects. Third, for the target subject, a subject-specific projection matrix is estimated using only the target training split, while keeping the shared space fixed. Finally, both source and target test data are projected into the shared space and evaluated using the trained decoder.

Source-subject performance is computed by evaluating the model on the test data of each source subject using their corresponding projection matrices. Target-subject performance is computed by applying the trained decoder to the held-out subject after alignment. This allows direct comparison between in-distribution and out-of-distribution performance under identical conditions.

\subsubsection{Baseline Methods}

To isolate the effect of shared-space alignment, we compare three different dimensionality reduction and alignment methods while keeping all other training settings identical: SRM-based, principal component analysis (PCA), and PCA combined with hyperalignment \cite{haxby2011common, haxby2020hyperalignment} (PCA+HA). This comparison is designed to disentangle the contribution of shared latent structure learning from that of dimensionality reduction and explicit cross-subject alignment. In particular, while HA focuses on aligning high-dimensional subject-specific representations into a common space by preserving representational geometry, SRM directly learns a low-dimensional shared latent representation that jointly captures shared structure across subjects. Therefore, this comparison allows us to evaluate whether explicitly modeling shared latent structure provides additional benefits for cross-subject semantic decoding beyond simple dimensionality reduction or alignment alone.

\subsubsection{Evaluation Metrics and Statistical Analysis}

We evaluate decoding performance in the semantic embedding space using word-level AUC-ROC, pairwise accuracy, and Top-$k$ accuracy. Statistical significance of performance differences between decoding paradigms was assessed using Wilcoxon signed-rank tests \cite{wilcoxon1945individual}, with significance defined as $p < 0.01$. Word-level AUC-ROC treats each word as a class and evaluates how well samples belonging to a given word are distinguished from samples of all other words. For each word class, a class score is computed based on the average similarity between the predicted embedding and embeddings of samples belonging to that class. Using these scores, a one-versus-rest classification is performed for each word, and the results are averaged across all classes. Pairwise accuracy measures whether, for each sample, the predicted embedding is more similar to its corresponding ground-truth embedding than to mismatched ground-truth embeddings. In our implementation, the comparison items were not randomly sampled. Instead, all non-matching samples within the same evaluation mini-batch served as comparison items. The final pairwise accuracy was computed by averaging these all-pair comparisons across evaluation mini-batches. Top-$k$ accuracy evaluates whether the ground-truth word is included among the top $k$ predicted word classes for each sample. The candidate set consists of all word classes present in the test set. 

\section{Results}

\subsection{Decoding Performance}

Tables~I and~II summarize the decoding performance of the proposed SRM-based framework, PCA+HA, and the PCA baseline in the source-subject and held-out-subject settings, respectively. Across both evaluation settings and all metrics, SRM consistently achieves the highest performance, followed by PCA+HA, while PCA shows the lowest performance. As shown in Table~I, SRM yields the best source-subject performance across all metrics (AUC-ROC: $0.671 \pm 0.007$, pairwise accuracy: $0.667 \pm 0.007$, Top-10: $0.055 \pm 0.004$, Top-50: $0.222 \pm 0.007$). As shown in Table~II, SRM also maintains the strongest performance in the held-out-subject setting (AUC-ROC: $0.662 \pm 0.044$, pairwise accuracy: $0.656 \pm 0.035$, Top-10: $0.051 \pm 0.016$, Top-50: $0.209 \pm 0.039$). These results suggest that jointly learning a shared latent space across subjects is more effective than applying post hoc alignment after subject-wise dimensionality reduction.

\subsection{Cross-Subject Generalization}

To evaluate cross-subject generalization, we compare AUC-ROC performance between source-subject and held-out-subject settings, as shown in Fig.~3. SRM shows a modest performance decrease from source to held-out (AUC-ROC: $0.671 \pm 0.007$ to $0.662 \pm 0.044$), and this difference is not statistically significant ($p = 0.84375$), indicating stable generalization to held-out subjects. In contrast, PCA exhibits a large performance drop (AUC-ROC: $0.637 \pm 0.005$ to $0.553 \pm 0.029$), and this degradation is statistically significant ($p = 0.00781$), demonstrating inferior generalization performance across subjects. PCA+HA reduces this gap (AUC-ROC: $0.656 \pm 0.006$ to $0.640 \pm 0.040$), but still shows a noticeable decrease, although the difference is not statistically significant ($p = 0.54688$).

However, in the held-out condition, the SRM-based method ($0.662 \pm 0.044$) outperforms PCA+HA \mbox{($0.640 \pm 0.040$)}, and this difference is statistically significant \mbox{($p = 0.00781$)}, indicating that SRM provides more reliable performance on held-out subjects. Overall, the results show that SRM achieves the smallest performance gap between source and held-out settings and maintains statistically consistent performance, supporting its effectiveness in learning subject-invariant representations for cross-subject generalization.

\begin{figure}[t]
    \centering
    \includegraphics[width=\columnwidth]{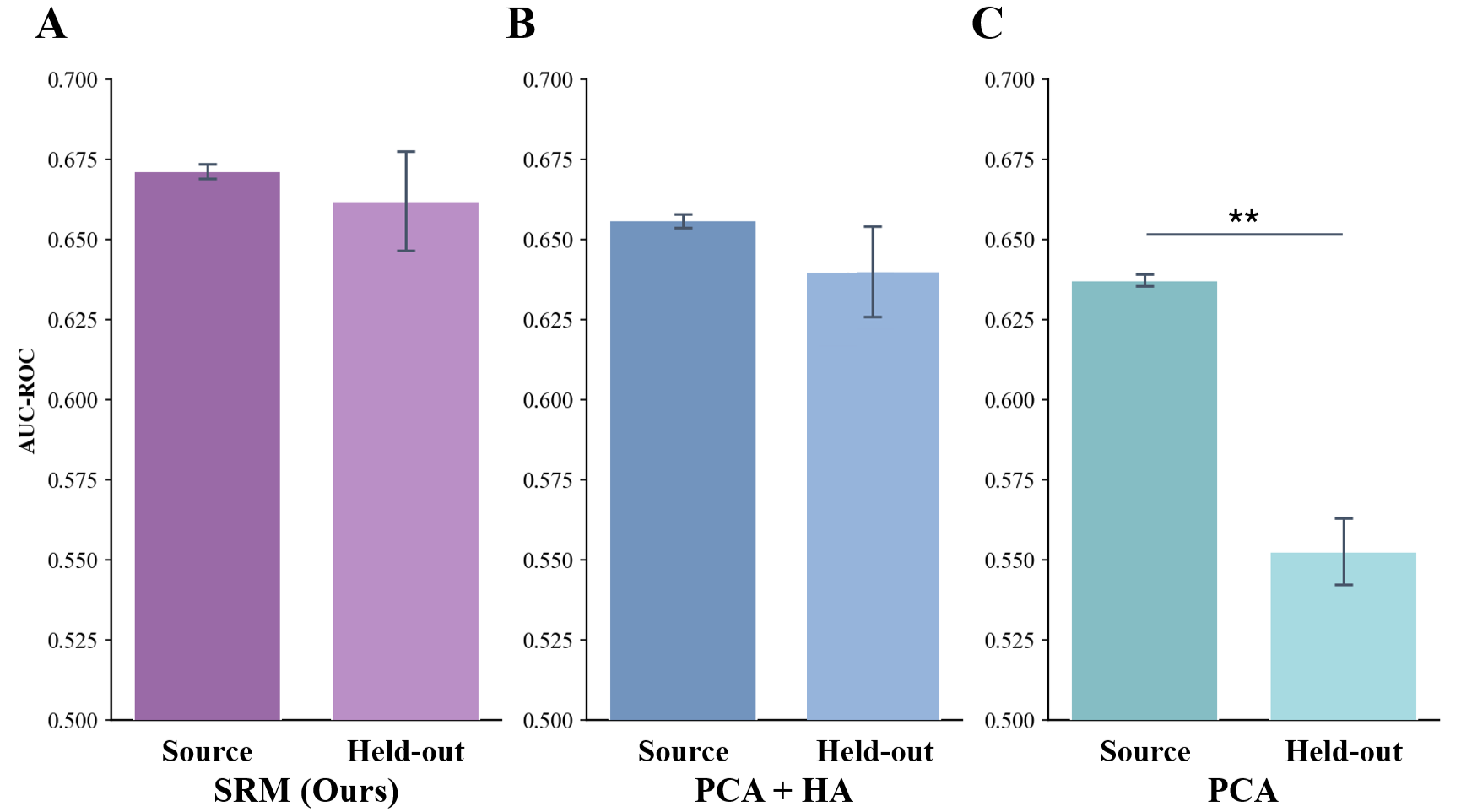}
    \caption{\textbf{Cross-subject generalization performance (AUC-ROC).}
    Decoding performance in source-subject and held-out-subject settings for (a) SRM, (b) PCA+HA, and (c) PCA. Among the three methods, SRM exhibits the smallest performance drop from source to held-out subjects, whereas PCA shows the largest degradation. PCA+HA lies between SRM and PCA, suggesting partial mitigation of cross-subject performance loss. A significant difference between source and held-out performance is observed for PCA (**$p < 0.01$). Error bars denote standard deviation across experiments.}
    \label{fig:barplot}
\end{figure}

\section{Discussion}

\subsection{Semantic Alignment for Cross-Subject Transferability}

The results indicate that the alignment strategy for cross-subject neural response signals is closely related to cross-subject generalization performance. In particular, SRM shows a small and statistically insignificant difference between the source and held-out settings, whereas PCA exhibits a significant performance degradation. These findings suggest that simple dimensionality reduction or post hoc alignment alone is insufficient to fully account for inter-subject variability. In the case of PCA+HA, PCA first preserves the dominant variance of each subject before alignment is performed. While this approach improves performance compared to PCA alone, it remains inferior to SRM overall and still shows performance degradation in the held-out setting, suggesting that subject-specific components retained prior to alignment are not fully removed.

Unlike PCA-based approaches, which preserve dominant within-subject variance, SRM emphasizes stimulus-locked components that are shared across individuals~\cite{chen2015reduced,bhattacharjee2026aligning}. As a result, it reduces subject-specific measurement variability arising from differences in electrode placement, signal scaling, and anatomical factors, while preserving task-relevant neural activity. These results suggest that effective cross-subject generalization depends on isolating neural responses that are consistently evoked by the same stimulus across individuals, rather than maximizing variance within each subject.

From a neurophysiological perspective, this shared latent space can be interpreted as reflecting common neural response structure associated with language processing across subjects. Neural activity during speech perception is distributed across multiple cortical regions, and similar stimuli are known to evoke reproducible, stimulus-locked responses despite differences in recording configurations \cite{goldstein2022shared, huth2016natural}. The ability to align multiple subjects into a common space therefore suggests that such shared response structure is sufficiently consistent to be captured and adopted for decoding.

\subsection{Implications in Semantic Decoding}

The results also suggest that the alignment strategy for cross-subject neural response may influence semantic decoding. SRM outperforms PCA+HA across both ranking-based and retrieval-based metrics, indicating that the learned representations are stable across different evaluation criteria. In PCA-based approaches, subject-specific variability is partially retained due to the nature of the method, which may require the decoder to account for both semantic information and irrelevant variability. In contrast, SRM suppresses such variability and aligns neural representations corresponding to similar semantic content into a consistent structure across subjects \cite{bhattacharjee2026aligning}, thereby providing a more stable input to the decoder. This alignment likely reduces variability in the mapping from neural response signals to the shared space, and may contribute to improved semantic decoding performance.

Importantly, these interpretations can be considered in the context of a perceived speech task using podcast data. In this setting, semantic content is clearly defined and neural responses are relatively well-aligned across subjects in time, providing favorable conditions for learning shared representations \cite{zada2025podcast}. In more challenging scenarios, such as imagined or attempted speech, neural signals may exhibit greater variability both within and across subjects than in settings where the same external stimulus is presented to all subjects, which may make it more difficult to construct a stable shared latent space \cite{wandelt2024representation}. Nevertheless, it would be an interesting direction for future research to investigate whether the shared-space alignment method that contributed to improved generalization performance in this study can provide similar generalization benefits for models that decode neural signals obtained from such more complex and realistic tasks.


\section{CONCLUSION}

In this study, we proposed a cross-subject semantic decoding framework based on shared-space alignment using SRM. Beyond prior SRM-based approaches that focus on encoding models, our framework demonstrates that SRM can also be effectively applied to semantic decoding tasks that reconstruct meaning from neural response signals.
Experimental results show that SRM-based alignment improves decoding performance while mitigating the performance drop observed when the model is applied to held-out subjects, thereby enabling more stable generalization. These findings suggest that shared neural representations may play an important role for achieving robust cross-subject semantic decoding, and support the potential of shared-space alignment as a scalable direction for developing generalized neural decoding models with minimal subject-specific adaptation.

\addtolength{\textheight}{-12cm}   



\bibliographystyle{IEEEtran}
\bibliography{REFERENCE}

\end{document}